\documentclass[10pt,twocolumn,letterpaper]{article}

\usepackage[pagenumbers]{cvpr} 

\usepackage{graphicx}
\usepackage{amsmath}
\usepackage{amssymb}
\usepackage{booktabs}

\usepackage{multirow}
\usepackage{colortbl} 
\usepackage{xcolor} 

\usepackage{animate}

\usepackage[pagebackref,breaklinks,colorlinks]{hyperref}

\usepackage[capitalize]{cleveref}
\crefname{section}{Sec.}{Secs.}
\Crefname{section}{Section}{Sections}
\Crefname{table}{Table}{Tables}
\crefname{table}{Tab.}{Tabs.}

\newcommand{\bx}{{\mathbf{x}}}

\newcommand{\bz}{{\mathbf{z}}}
\newcommand{\bI}{{\mathbf{I}}}
\newcommand{\bV}{{\mathbf{V}}}
\newcommand{\bW}{{\mathbf{W}}}
\newcommand{\bX}{{\mathbf{X}}}
\newcommand{\bY}{{\mathbf{Y}}}
\newcommand{\bZ}{{\mathbf{Z}}}
\newcommand{\bbR}{{\mathbb{R}}}
\newcommand{\comment}[1]{}

\def\cvprPaperID{6849} 
\def\confName{CVPR}
\def\confYear{2022}

\begin{document}

\title{Capturing Humans in Motion: Temporal-Attentive 3D Human Pose and Shape Estimation from Monocular Video\vspace{-12pt}}


\author{Wen-Li Wei$^*$, Jen-Chun Lin$^*$, Tyng-Luh Liu, and Hong-Yuan Mark Liao$^{\dag}$\\
Institute of Information Science, Academia Sinica, Taiwan\\
}

\twocolumn[{%
\renewcommand\twocolumn[1][]{#1}%
\maketitle
\begin{center}
    \centering
    \captionsetup{type=figure}\vspace{-25pt}
    \includegraphics[width=.92\textwidth]{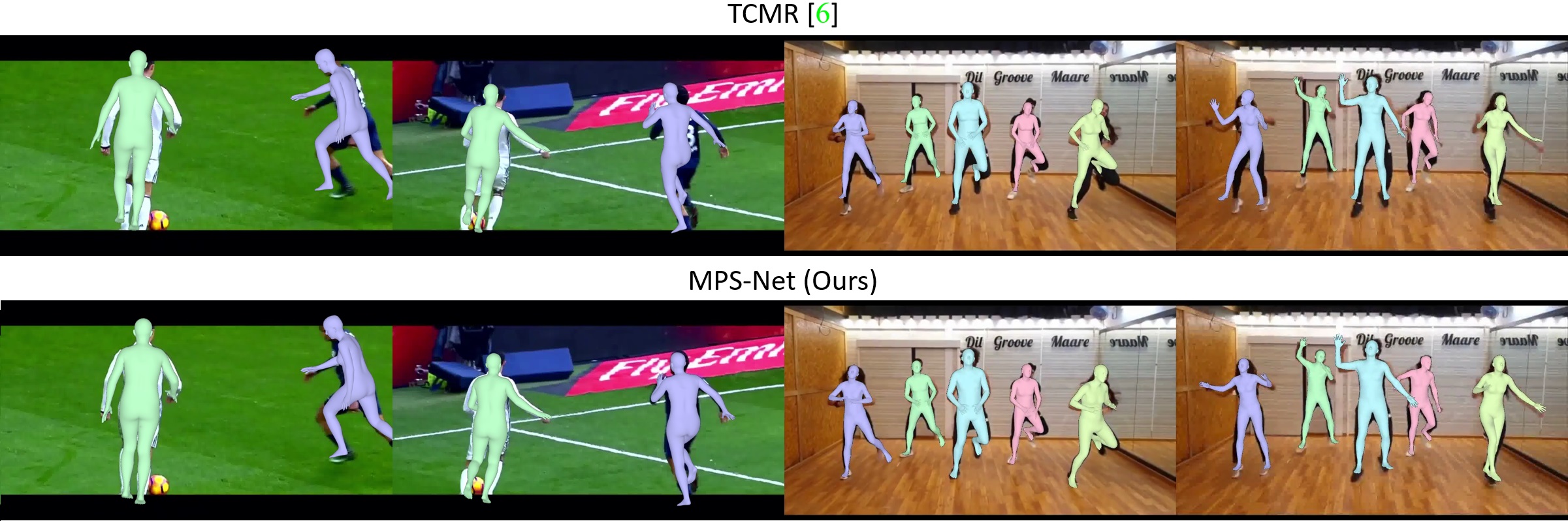}\vspace{-8pt}
    \caption{By coupling \emph{motion continuity attention} with \emph{hierarchical attentive feature integration}, the proposed MPS-Net can achieve more accurate pose and shape estimations (bottom row), when dealing with in-the-wild videos. For comparison, the results (top row) obtained by TCMR \cite{choi2020beyond}, the state-of-the-art video-based 3D human pose and shape estimation method, are included. }\vspace{-3pt}
    \label{fig:teaser}
\end{center}%
}]
\def\thefootnote{*}\footnotetext{Both authors contributed equally to this work}
\def\thefootnote{\dag}\footnotetext{Mark Liao is also a Chair Professor of Providence University}

\begin{abstract}
   \vspace{-13pt}Learning to capture human motion is essential to 3D human pose and shape estimation from monocular video. However, the existing methods mainly rely on recurrent or convolutional operation to model such temporal information, which limits the ability to capture non-local context relations of human motion. To address this problem, we propose a motion pose and shape network (MPS-Net) to effectively capture humans in motion to estimate accurate and temporally coherent 3D human pose and shape from a video. Specifically, we first propose a motion continuity attention (MoCA) module that leverages visual cues observed from human motion to adaptively recalibrate the range that needs attention in the sequence to better capture the motion continuity dependencies. Then, we develop a hierarchical attentive feature integration (HAFI) module to effectively combine adjacent past and future feature representations to strengthen temporal correlation and refine the feature representation of the current frame. By coupling the MoCA and HAFI modules, the proposed MPS-Net excels in estimating 3D human pose and shape in the video. Though conceptually simple, our MPS-Net not only outperforms the state-of-the-art methods on the 3DPW, MPI-INF-3DHP, and Human3.6M benchmark datasets, but also uses fewer network parameters. The video demos can be found at \url{https://mps-net.github.io/MPS-Net/}.
\end{abstract}
\vspace{-12pt}
\section{Introduction}\vspace{-7pt}
Estimating 3D human pose and shape by taking a simple picture/video without relying on sophisticated 3D scanning devices or multi-view stereo algorithms, has important applications in computer graphics, AR/VR, physical therapy and beyond. Generally speaking, the task is to take a single image or video sequence as input and to estimate the parameters of a 3D human mesh model as output. Take, for example, the SMPL model \cite{Loper2015SMPLAS}. For each image, it needs to estimate $85$ (including pose, shape, and camera) parameters, which control the $6890$ vertices that form the full 3D mesh of a human body \cite{Loper2015SMPLAS}. Despite recent progress on 3D human pose and shape estimation, it is still a frontier challenge due to depth ambiguity, limited 3D annotations, and complex motion of non-rigid human body \cite{hmrKanazawa17,Kolotouros2019LearningTR,Kocabas2020VIBEVI,choi2020beyond}.

Different from 3D human pose and shape estimation from a single image \cite{hmrKanazawa17,omran2018nbf,Pavlakos2018LearningTE,Kolotouros2019LearningTR,Georgakis2020HierarchicalKH}, estimating it from monocular video is a more complex task \cite{Kanazawa2019Learning3H,Doersch2019Sim2realTL,Yu2019HumanMR,Kocabas2020VIBEVI,Luo_2020_ACCV,choi2020beyond}. It needs to not only estimate the pose, shape and camera parameters of each image, but also correlate the continuity of human motion in the sequence. Although existing single image-based methods can predict a reasonable output from a static image, it is difficult for them to estimate temporally coherent and smooth 3D human pose and shape in the video sequence due to the lack of modeling the continuity of human motion in consecutive frames.
To solve this problem, several methods have recently been proposed to extend the single image-based methods to the video cases, which mainly rely on recurrent neural network (RNN) or convolutional neural network (CNN) to model temporal information (\ie continuity of human motion) for coherent predictions \cite{Kanazawa2019Learning3H,Doersch2019Sim2realTL,Kocabas2020VIBEVI,Luo_2020_ACCV,choi2020beyond}. However, RNNs and CNNs are good at dealing with local neighborhoods \cite{Vaswani2017AttentionIA,Wang2018NonlocalNN}, and the models alone may not be effective for learning long-range dependencies (\ie non-local context relations) between feature representations to describe the relevance of human motion. As a result, there is still room for improvement for existing video-based methods to estimate accurate and smooth 3D human pose and shape (see Figure \ref{fig:teaser}).

\begin{figure}
  \centering
  \includegraphics[width=.96\linewidth]{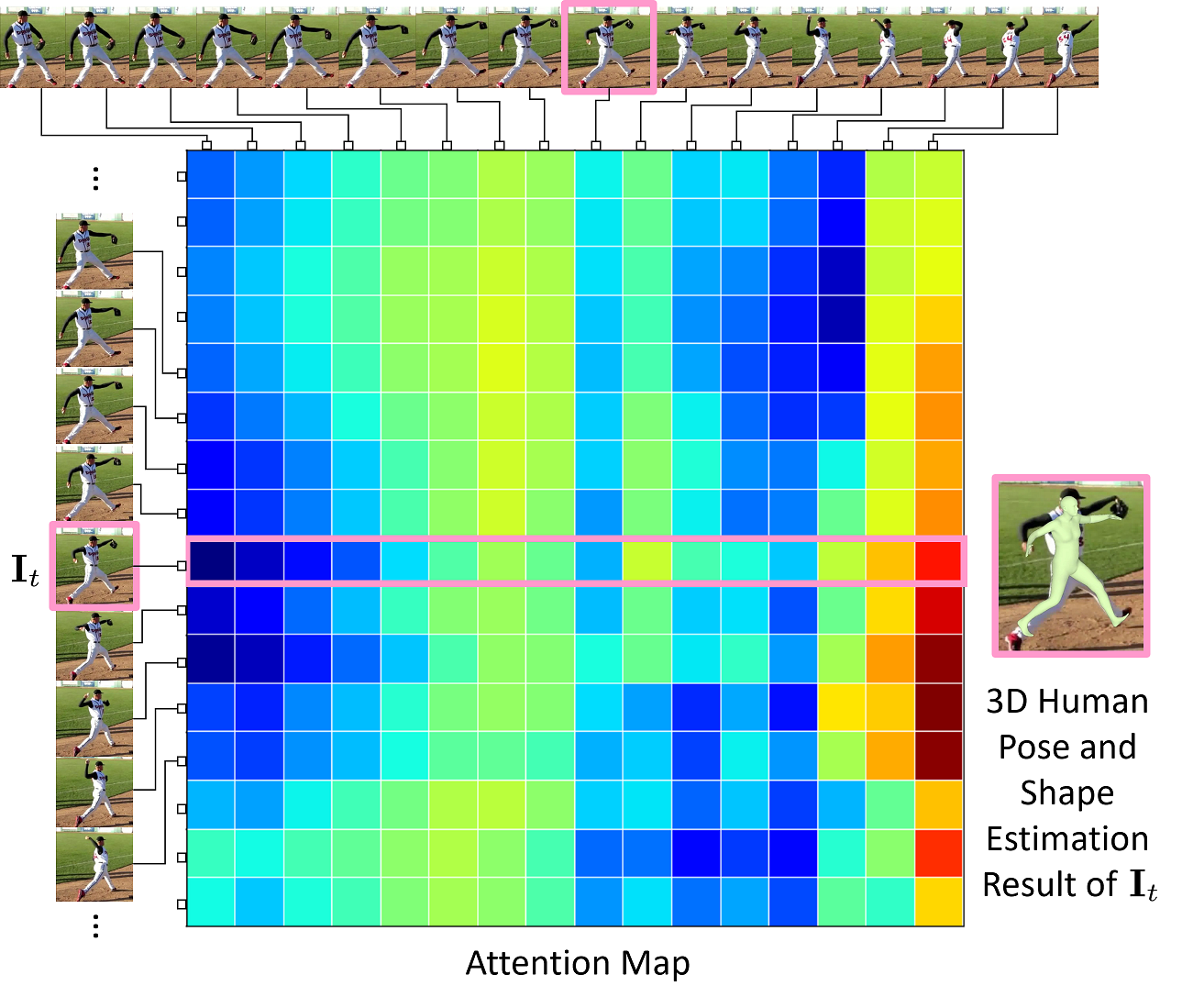}\vspace{-13pt}
  \caption{Visualization of the attention map generated by the self-attention module \cite{Wang2018NonlocalNN} in 3D human pose and shape estimation. The visualization shows that the attention map is easy to focus attention on less correlated temporal positions (\ie far apart frames with very different action poses) and lead to inaccurate 3D human pose and shape estimation (see frame $\bI_t$). In the attention map, red indicates a higher attention value, and blue indicates a lower one.}\vspace{-13pt}
  \label{fig:fig2}
\end{figure}

To address the aforementioned issue, we propose a motion pose and shape network (MPS-Net) for 3D human pose and shape estimation from monocular video. Our key insights are two-fold. First, although a self-attention mechanism \cite{Vaswani2017AttentionIA,Wang2018NonlocalNN} has recently been proposed to compensate (\ie better learn long-range dependencies) for the weaknesses of recurrent and convolutional operations, we empirically find that it is not always good at modeling human motion in the action sequence. Because the attention map computed by the self-attention module is often unstable, which is easy to focus attention on less correlated temporal positions (\ie far apart frames with very different action poses) and ignore the continuity of human motion in the action sequence (see Figure \ref{fig:fig2}). To this end, we propose a motion continuity attention (MoCA) module to achieve the adaptability to diverse temporal content and relations in the action sequence. Specifically, the MoCA module contributes in two points. First, a normalized self-similarity matrix (NSSM) is developed to capture the structure of temporal similarities and dissimilarities of visual representations in the action sequence, thereby revealing the continuity of human motion. Second, NSSM is regarded as the a \emph{priori} knowledge and applied to guide the learning of the self-attention module, which allows it to adaptively recalibrate the range that needs attention in the sequence to capture the motion continuity dependencies.
In the second insight, motivated by the temporal feature integration scheme in 3D human mesh estimation \cite{choi2020beyond}, we develop a hierarchical attentive feature integration (HAFI) module that utilizes adjacent feature representations observed from past and future frames to strengthen temporal correlation and refine the feature representation of the current frame. By coupling the MoCA and HAFI modules, our MPS-Net can effectively capture humans in motion to estimate accurate and temporally coherent 3D human pose and shape from monocular video (see Figure \ref{fig:teaser}). We characterize the main contributions of our MPS-Net as follows:\vspace{-9pt}

\begin{itemize}
\item We propose a MoCA module that leverages visual cues observed from human motion to adaptively recalibrate the range that needs attention in the sequence to better capture the motion continuity dependencies.\vspace{-6pt}
\item We develop a HAFI module that effectively combines adjacent past and future feature representations in a hierarchical attentive integration manner to strengthen temporal correlation and refine the feature representation of the current frame.\vspace{-7pt}
\item Extensive experiments on three standard benchmark datasets demonstrate that our MPS-Net achieves the state-of-the-art performance against existing methods and uses fewer network parameters.\vspace{-6pt}
\end{itemize}

\section{Related work}\vspace{-5pt}
\label{sec:formatting}

{\bf 3D human pose and shape estimation from a single image.} The existing single image-based 3D human pose and shape estimation methods are mainly based on parametric 3D human mesh models, such as SMPL \cite{Loper2015SMPLAS}, \ie trains a deep-net model to estimate pose, shape, and camera parameters from the input image, and then decodes them into a 3D mesh of the human body through the SMPL model. For example, Kanazawa \etal \cite{hmrKanazawa17} proposed an end-to-end human mesh recovery (HMR) framework to regress SMPL parameters from a single RGB image. They employ 3D to 2D keypoint reprojection loss and adversarial training to alleviate the limited 3D annotation problem and make the output 3D human mesh anatomically reasonable. Pavlakos \etal \cite{Pavlakos2018LearningTE} used 2D joint heatmaps and silhouette as cues to improve the accuracy of SMPL parameter estimation. Similarly, Omran \etal \cite{omran2018nbf} used a semantic segmentation scheme to extract body part information as a cue to estimate the SMPL parameters. Kolotouros \etal \cite{Kolotouros2019LearningTR} proposed a self-improving framework that integrates the SMPL parameter regressor and iterative fitting scheme to better estimate 3D human pose and shape. Zhang \etal \cite{pymaf2021} designed a pyramidal mesh alignment feedback (PyMAF) loop in the deep SMPL parameter regressor to exploit multi-scale contexts for better mesh-image alignment of the reconstruction.\vspace{-2pt} 

Several non-parametric 3D human mesh reconstruction methods \cite{Kolotouros2019ConvolutionalMR,varol18_bodynet,Moon_2020_ECCV_I2L-MeshNet} have been proposed. For example, Kolotouros \etal \cite{Kolotouros2019ConvolutionalMR} proposed a graph CNN, which takes the 3D human mesh template and image embedding (extracted from ResNet-50 \cite{He2016DeepRL}) as input to directly regress the vertex coordinates of the 3D mesh. Moon and Lee \cite{Moon_2020_ECCV_I2L-MeshNet} proposed an I2L-MeshNet, which uses a lixel-based 1D heatmap to directly localize the vertex coordinates of the 3D mesh in a fully convolutional manner.\vspace{-2pt}


Despite the above methods are effective for static images, they are difficult to generate temporally coherent and smooth 3D human pose and shape in the video sequence, \ie jittery, unstable 3D human motion may occur \cite{Kocabas2020VIBEVI,choi2020beyond}.\vspace{3pt}

{\bf 3D human pose and shape estimation from monocular video.} Similar to the single image-based methods, the existing video-based 3D human pose and shape estimation methods are mainly based on the SMPL model. For example, Kanazawa \etal \cite{Kanazawa2019Learning3H} proposed a convolution-based temporal encoder to learn human motion kinematics by further estimating SMPL parameters in adjacent past and future frames. 
Doersch \etal \cite{Doersch2019Sim2realTL} trained their model on a sequence of 2D keypoint heatmaps and optical flow by combining CNN and long short-term memory (LSTM) network to demonstrate that considering pre-processed motion information can improve SMPL parameter estimation. Sun \etal \cite{Yu2019HumanMR} proposed a skeleton-disentangling framework, which divides the task into multi-level spatial and temporal sub-problems. They further proposed an unsupervised adversarial training strategy, namely temporal shuffles and order recovery, to encourage temporal feature learning. Kocabas \etal \cite{Kocabas2020VIBEVI} proposed a temporal encoder composed of bidirectional gated recurrent units (GRU) to encode static features into a series of temporally correlated latent features, and feed them to the regressor to estimate SMPL parameters. They further integrated adversarial training strategy that leverages the AMASS dataset \cite{Mahmood2019AMASSAO} to distinguish between real human motion and those estimated by its regressor to encourage the generation of reasonable 3D human motion. Luo \etal \cite{Luo_2020_ACCV} proposed a two-stage model that first estimates the coarse 3D human motion through a variational motion estimator, and then uses a motion residual regressor to refine the motion estimates. Recently, Choi \etal \cite{choi2020beyond} proposed a temporally consistent mesh recovery (TCMR) system that uses GRU-based temporal encoders with three different encoding strategies to encourage the network to better learn temporal features. In addition, they proposed a temporal feature integration scheme that combines the output of three temporal encoders to help the SMPL parameter regressor estimate accurate and smooth 3D human pose and shape.\vspace{-1pt}


Despite the success of RNNs and CNNs, both recurrent and convolutional operations can only deal with local neighborhoods \cite{Vaswani2017AttentionIA ,Wang2018NonlocalNN}, which makes it difficult for them to learn long-range dependencies (\ie non-local context relations) between feature representations in the action sequence. Therefore, existing methods are still struggling to estimate accurate and smooth 3D human pose and shape.\vspace{5pt}

{\bf Attention mechanism.} The attention mechanism has enjoyed widespread adoption as a computational module for natural language processing \cite{Vaswani2017AttentionIA,Devlin2019BERTPO,Radford2018ImprovingLU,Bahdanau2015NeuralMT,Yu2018QANetCL} and vision-related tasks \cite{Wang2018NonlocalNN,Hsieh2019OneShotOD,Dosovitskiy2020AnII,Hu2020SqueezeandExcitationN,Woo2018CBAMCB,Roy2019RecalibratingFC,Chen_2021_CVPR} because of its ability to capture long-range dependencies and selectively concentrate on the relevant subset of the input. There are various ways to implement the attention mechanism. Here we focus on self-attention \cite{Vaswani2017AttentionIA,Wang2018NonlocalNN}. For example, Vaswani \etal \cite{Vaswani2017AttentionIA} proposed a self-attention-based architecture called \emph{Transformer}, in which the self-attention module is designed to update each sentence’s element through the entire sentence’s aggregated information to draw global dependencies between input and output. The \emph{Transformer} entirely replaces the recurrent operation with the self-attention module, and greatly improves the performance of machine translation. Later, Wang \etal \cite{Wang2018NonlocalNN} showed that self-attention is an instantiation of non-local mean \cite{Buades2005ANA}, and proposed a non-local block for the CNN to capture long-range dependencies. Like the self-attention module proposed in \emph{Transformer}, the non-local operation computes the correlation between each position in the input feature representation to generate an attention map, and then performs the attention-guided dense context information aggregation to draw long-range dependencies.

Despite the self-attention mechanism performs well, we empirically find that the attention map computed by the self-attention module (\eg non-local block) is often unstable, which means that it is easy to focus attention on less correlated temporal positions (\ie far apart frames with very different action poses) and ignore the continuity of human motion in the action sequence (see Figure \ref{fig:fig2}). In this work, we propose the MoCA module, which extends the learning of the self-attention module by introducing the a \emph{priori} knowledge of NSSM to adaptively recalibrate the range that needs attention in the sequence, so as to capture motion continuity dependencies. The HAFI module is further proposed to strengthen the temporal correlation and refine the feature representation of each frame through its neighbors.\vspace{-1pt}


\vspace{-3pt}
\section{Method}
\begin{figure*}[t]
  \centering
  \includegraphics[width=.92\linewidth]{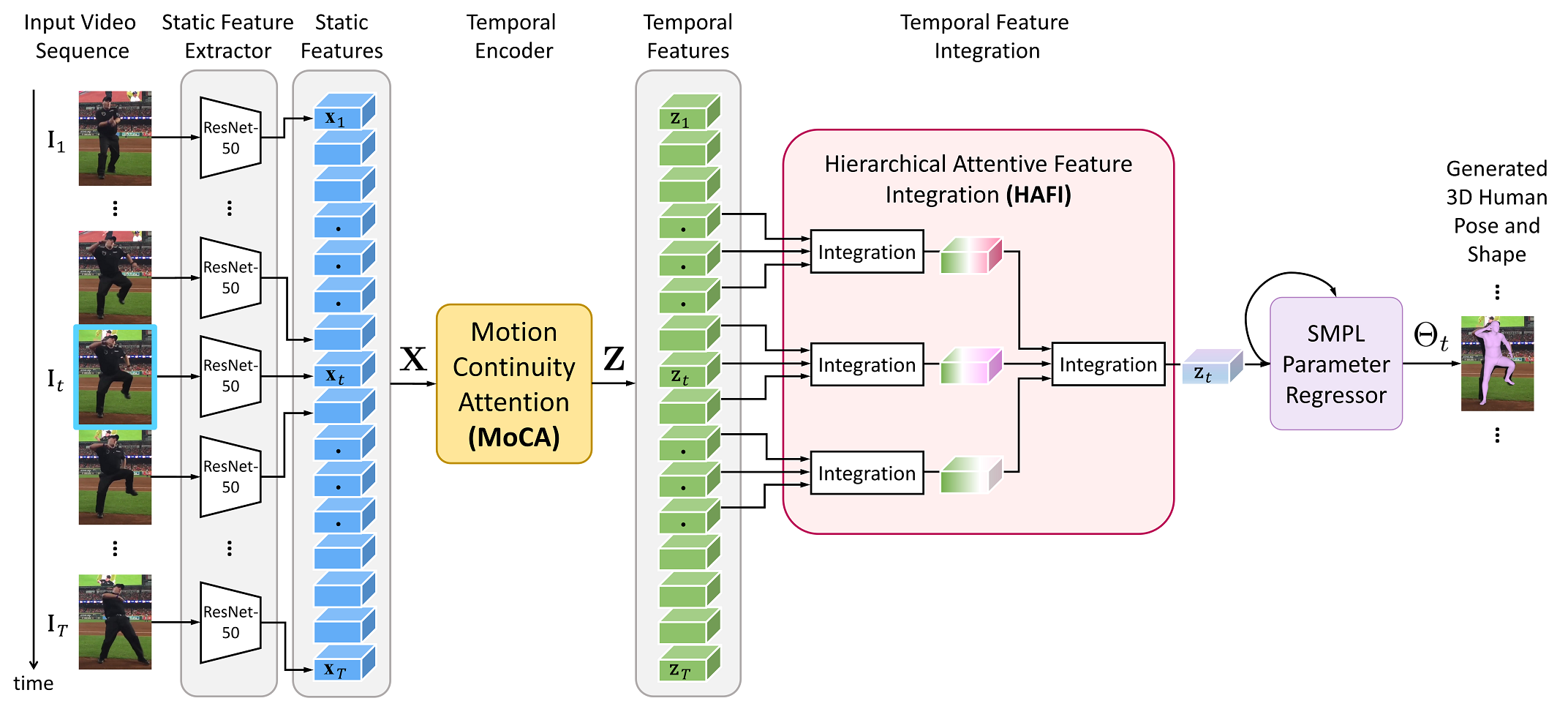}\vspace{-11pt}
  \caption{Overview of our motion pose and shape network (MPS-Net). MPS-Net estimates pose, shape, and camera parameters $\Theta$ in the video sequence based on the static feature extractor, temporal encoder, temporal feature integration, and SMPL parameter regressor to generate 3D human pose and shape.}\vspace{-13pt}
  \label{fig:fig3}
\end{figure*}

\vspace{-3pt}
Figure~\ref{fig:fig3} shows the overall pipeline of our MPS-Net. We elaborate each module in MPS-Net as follows.\vspace{-2pt}

\begin{figure}
  \centering
  \includegraphics[width=.81\linewidth]{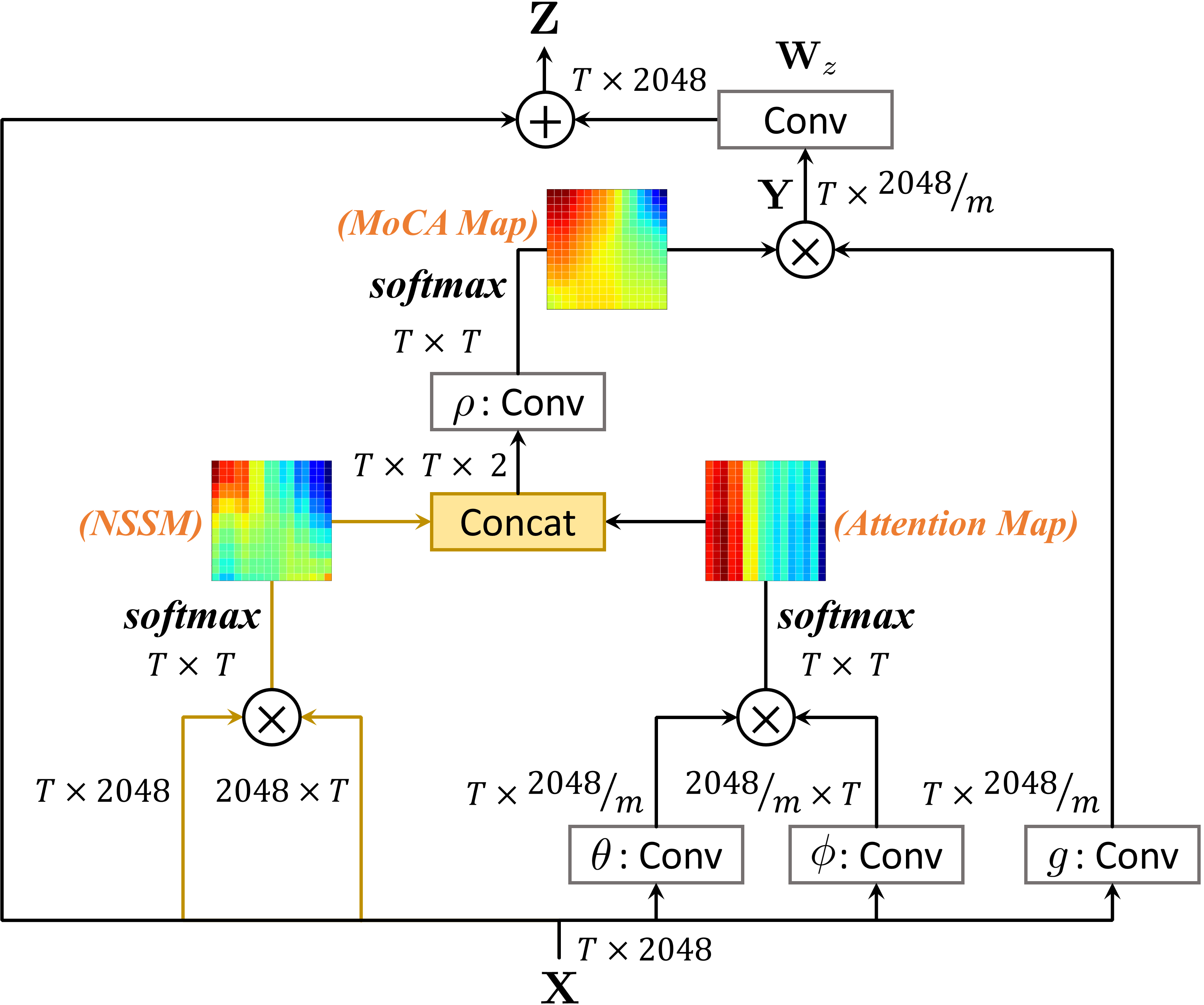}\vspace{-7pt}
  \caption{A MoCA module. $\mathbf{X}$ is shown as the shape of $T \times 2048$ for $2048$ channels. $g$, $\phi$, $\theta$, and $\rho$ denote convolutional operations, $\otimes$ denotes matrix multiplication, and $\oplus$ denotes element-wise sum. The computation of softmax is performed on each row.}\vspace{-13pt}
  \label{fig:fig4}
\end{figure}

\vspace{-3pt}
\subsection{Temporal encoder}\vspace{-3pt}
Given an input video sequence $\bV=\{ \bI_t \}_{t=1}^T$ with $T$ frames. We first use ResNet-50 \cite{He2016DeepRL} pre-trained by Kolotouros \etal \cite{Kolotouros2019LearningTR} to extract the static feature of each frame to form a static feature representation sequence $\bX =\{\bx_t \}_{t=1}^T$, where $\bx_t \in \bbR^{2048}$. Then, the extracted $\bX$ is sent to the proposed MoCA module to calculate the temporal feature representation sequence $\bZ = \{ \bz_t \}_{t=1}^T$, where $\bz_t \in \bbR^{2048}$.\vspace{4pt}

\noindent{\bf MoCA Module.} We propose a MoCA operation to extend the non-local operation \cite{Wang2018NonlocalNN} in two ways. First, we introduce an NSSM to capture the structure of temporal similarities and dissimilarities of visual representations in the action sequence to reveal the continuity of human motion. Second, we regard NSSM as the a \emph{priori} knowledge and combine it with the attention map generated by the non-local operation to adaptively recalibrate the range that needs attention in the action sequence.\vspace{-2pt}



We formulate the proposed MoCA module as follows (see Figure \ref{fig:fig4}). Given the static feature representation sequence $\bX \in \bbR^{T \times 2048}$, the goal of the MoCA operation is to obtain a non-local context response $\bY \in \bbR^{T \times \frac{2048}{m}}$, which aims to capture the \textit{motion continuity dependencies} across the whole representation sequence by weighted sum of the static features at all temporal positions,\vspace{-4pt}
\begin{equation}\label{eq:MCSAmodule1}
\bY = \rho(\lbrack f(\bX, \bX), f(\theta(\bX), \phi(\bX)) \rbrack) g(\bX),\vspace{-3pt}
\end{equation}
where $m$ is a reduction ratio used to reduce computational complexity \cite{Wang2018NonlocalNN}, and it is set to $2$ in our experiments. $g(\cdot)$, $\phi(\cdot)$, and  $\theta(\cdot)$ are learnable transformations, which are implemented by using the convolutional operation \cite{Wang2018NonlocalNN}. Thus, the transformations can be written as\vspace{-4pt}
\begin{equation}\label{eqn:NLoperation4}
g(\bX)=\bX\bW_g \in \bbR^{T \times \frac{2048}{m}},\vspace{-4pt}
\end{equation} 
\begin{equation}\label{eqn:NLoperation3}
\phi(\bX)=\bX\bW_\phi \in \bbR^{T \times \frac{2048}{m}},\vspace{-4pt}
\end{equation}
and\vspace{-5pt}
\begin{equation}\label{eqn:NLoperation2}
\theta(\bX)=\bX\bW_\theta \in \bbR^{T \times \frac{2048}{m}},
\end{equation}
parameterized by the weight matrices $\bW_g$,
$\bW_\phi$, and
$\bW_\theta\in \bbR^{2048 \times \frac{2048}{m}}$, respectively. $f(\cdot,\cdot)$ represents a pairwise function, which computes the affinity between all positions. We use dot product \cite{Wang2018NonlocalNN} as the operation for $f$, \ie\vspace{-4pt}
\begin{equation}\label{eqn:NLoperation5}
f(\theta(\bX), \phi(\bX))= \theta(\bX) \phi(\bX)^\mathsf{T},\vspace{-4pt}
\end{equation}
where the size of the resulting pairwise function $f(\theta(\bX), \phi(\bX))$ is denoted as $\bbR^{T \times \frac{2048}{m}} \times \bbR^{\frac{2048}{m} \times T} \to \bbR^{T \times T}$, which encodes the mutual similarity between temporal positions under the transformed static feature representation sequence. Then, the softmax operation is used to normalize it into an attention map (see Figure \ref{fig:fig4}). 

\begin{figure*}[t]
  \centering
  \includegraphics[width=.93\linewidth]{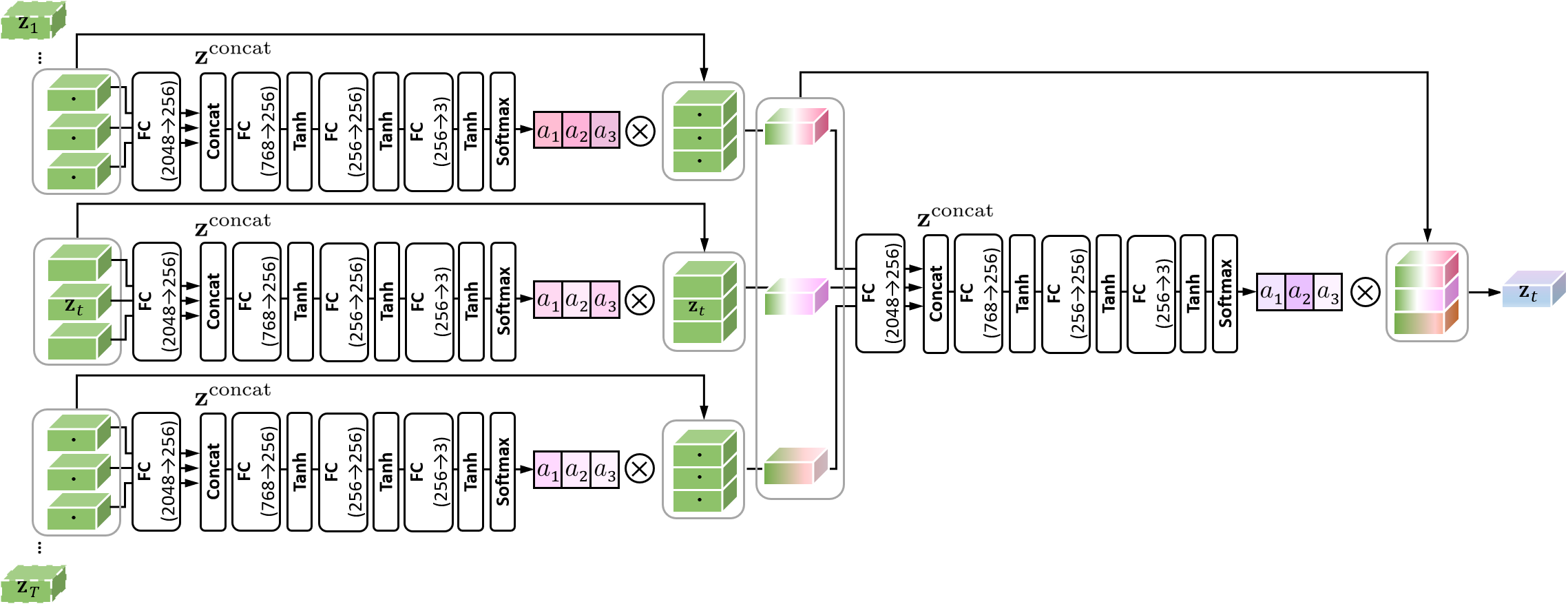}\vspace{-9pt}
  \caption{A HAFI module. It utilizes the temporal features observed from the past and future frames to refine the temporal feature of the current frame $\bz_t$ in a hierarchical attentive integration manner. Where $\otimes$ denotes matrix multiplication.}\vspace{-12pt}
  \label{fig:fig5}
\end{figure*}

We empirically find that although calculating the similarity in the transformed feature space provides an opportunity for insight into implicit long-range dependencies, it may sometimes be unstable and lead to attention on less correlated temporal positions (see Figure \ref{fig:fig2}). To this end, we introduce NSSM into the MoCA operation to enable the MoCA module to learn to focus attention on a more appropriate range of action sequence.


Regarding NSSM construction, unlike the non-local operation \cite{Wang2018NonlocalNN}, we directly use the static feature representation sequence $\bX$ extracted from the input video to reveal the explicit dependencies between the frames through the self-similarity matrix \cite{10.1145/319463.319472} construction $f(\bX, \bX)=\bX \bX^\mathsf{T} \in \bbR^{T \times T}$. In this way, the continuity of human motion in the input video can be more straightforwardly revealed. Similarly, we normalize the resultant self-similarity matrix through the softmax operation to form an NSSM (see Figure \ref{fig:fig4}) to facilitate subsequent combination with the attention map.

For the combination of NSSM and attention map, we first regard NSSM as the a \emph{priori} knowledge to concatenate the attention map through the operation $\lbrack \cdot,\cdot \rbrack$, and then use the learnable transformation $\rho(\cdot)$, \ie $1\times1$ convolution to recalibrate the attention map by referring to NSSM (see Figure \ref{fig:fig4} and \cref{eq:MCSAmodule1}). The resultant $\rho(\cdot)$ is then normalized through the softmax operation, which is called the MoCA map. By jointly considering the characteristics of the NSSM and the attention map, the MoCA map can reveal the non-local context relations related to the human motion of the input video in a more appropriate range. To this end, the non-local context response $\bY \in \bbR^{T \times \frac{2048}{m}}$ can be calculated from the linear combination between the matrices resulted from $\rho(\cdot)$ and $g(\cdot)$.


Finally, as in the design of the non-local block \cite{Wang2018NonlocalNN}, we use residual connection \cite{He2016DeepRL} to generate the output temporal feature representation sequence $\bZ \in \bbR^{T \times 2048}$ (see Figure~\ref{fig:fig4}) in the MoCA module as follows:\vspace{-6pt}
\begin{equation}\label{eqn:MCSAmodule2}
\bZ=\bY\bW_{z}+\bX,\vspace{-7pt}
\end{equation}
where $\bW_{z}$ is a learnable weight matrix implemented by using the convolutional operation \cite{Wang2018NonlocalNN}, and the number of channels in $\bW_{z}$ is scaled up to match the number of channels (\ie $2048$) in $\bX$. ``$+\bX$'' denotes a residual connection. The residual connection allows us to insert the MoCA module into any pre-trained network, without breaking its initial behavior (\eg if $\bW_{z}$ is initialized as zero). As a result, by further considering the non-local context response $\bY$, $\bZ$ will contain rich temporal information, so $\bZ$ can be regarded as enhanced $\bX$.\vspace{-4pt}

\subsection{Temporal feature integration}\vspace{-5pt}

Given the temporal feature representation sequence $\bZ \in \bbR^{T \times 2048}$, the goal of the HAFI module is to refine the temporal feature of the current frame $\bz_t$ by integrating the adjacent temporal features observed from past and future frames to strengthen their temporal correlation and obtain better pose and shape estimation, as shown in Figure \ref{fig:fig3}.

\noindent{\bf HAFI Module.}
Specifically, we use $T/2$ adjacent frames (\ie $\{\bz_{t \pm \frac{T}{4}}\}$) to refine the temporal feature of the current frame $\bz_t$ in a hierarchical attentive integration manner, as shown in Figure \ref{fig:fig5}.
For each branch in the HAFI module, we consider the temporal features of three adjacent frames as a group (adjacent frames between groups do not overlap), and resize them from $2048$ dimensions to $256$ dimensions respectively through a shared fully connected (FC) layer to reduce computational complexity. The resized temporal features are concatenated ($\bz^{\mathrm{concat}} \in \bbR^{768}$) and passed to three FC layers and a softmax activation to calculate the attention values $\mathbf{a}=\lbrace a_k \rbrace_{k=1}^3$ by exploring the dependencies among them. Then, the attention value is weighted back to each corresponding frame to amplify the contribution of important frames in the temporal feature integration to obtain the aggregated temporal feature (see Figure \ref{fig:fig5}). The aggregated temporal features produced by the bottom branches will be passed to the upper layer and integrated in the same way to produce the final refined $\bz_t$. By gradually integrating temporal features in adjacent frames to strengthen temporal correlation, it will provide opportunities for the SMPL parameter regressor to learn to estimate accurate and temporally coherent 3D human pose and shape.

In this work, like Kocabas \etal \cite{Kocabas2020VIBEVI}, we use the SMPL parameter regressor proposed in \cite{hmrKanazawa17, Kolotouros2019LearningTR} as our regressor to estimate pose, shape, and camera parameters $\Theta_t \in \bbR^{85}$ according to each refined $\bz_t$ (see Figure~\ref{fig:fig3}). In the training phase, we initialize the SMPL parameter regressor with pre-trained weights from HMR \cite{hmrKanazawa17, Kolotouros2019LearningTR}.\vspace{-3pt}

\subsection{Loss functions}\vspace{-4pt}


In terms of MPS-Net training, for each estimated $\Theta_t$, following the method proposed by Kocabas \etal \cite{Kocabas2020VIBEVI}, we impose $\mathcal{L}_{\mathrm{2}}$ loss between the estimated and ground-truth SMPL parameters and 3D/2D joint coordinates to supervise MPS-Net to generate reasonable real-world poses. The 3D joint coordinates are obtained by forwarding the estimated SMPL parameters to the SMPL model \cite{Loper2015SMPLAS}, and the 2D joint coordinates are obtained through the 2D projection of the 3D joints using the predicted camera parameters \cite{Kocabas2020VIBEVI}. In addition, like Kocabas \etal \cite{Kocabas2020VIBEVI}, we also apply adversarial loss $\mathcal{L}_{\mathrm{adv}}$, \ie using the AMASS \cite{Mahmood2019AMASSAO} dataset to train a discriminator to distinguish between real human motion and those generated by MPS-Net's SMPL parameter regressor to encourage the generation of reasonable 3D human motion.\vspace{-6pt}

\begin{table*}
\fontsize{7.3pt}{10pt}\selectfont
\begin{center}
\begin{tabular}{l@{\hspace{0.13cm}} | c@{\hspace{0.13cm}}c@{\hspace{0.13cm}}c@{\hspace{0.13cm}}c@{\hspace{0.13cm}} | c@{\hspace{0.13cm}}c@{\hspace{0.13cm}}c@{\hspace{0.13cm}} | c@{\hspace{0.13cm}}c@{\hspace{0.13cm}}c@{\hspace{0.13cm}} | c@{\hspace{0.1cm}}}
\toprule 
&\multicolumn{4}{c}{3DPW}&\multicolumn{3}{c}{MPI-INF-3DHP}&\multicolumn{3}{c|}{Human3.6M} & Number of \\
\cmidrule(lr){2-5} \cmidrule(lr){6-8} \cmidrule(lr){9-11}
Method & PA-MPJPE $\downarrow$ & MPJPE $\downarrow$ & MPVPE $\downarrow$ & ACC-ERR  $\downarrow$ &PA-MPJPE $\downarrow$ & MPJPE $\downarrow$ & ACC-ERR  $\downarrow$ &PA-MPJPE $\downarrow$ & MPJPE $\downarrow$ & ACC-ERR  $\downarrow$ & Input Frames\\
\midrule
VIBE \cite{Kocabas2020VIBEVI} & 57.6 & 91.9 & - & 25.4 & 68.9 & 103.9 & 27.3 & 53.3 & 78.0 & 27.3 & \bf{16} \\
\rowcolor{black!10}MEVA \cite{Luo_2020_ACCV} & 54.7 & 86.9 & - & 11.6 & 65.4 & \bf{96.4} & 11.1 & 53.2 & 76.0 & 15.3 & 90 \\
TCMR \cite{choi2020beyond} & 52.7 & 86.5 & 103.2 & \bf{6.8} & 63.5 & 97.6 & \bf{8.5} & 52.0 & 73.6 & 3.9 & \bf{16} \\
\rowcolor{black!10}MPS-Net (Ours) & \bf{52.1} & \bf{84.3} & \bf{99.7} & 7.4 & \bf{62.8} & 96.7 & 9.6 & \bf{47.4} & \bf{69.4} & \bf{3.6} & \bf{16} \\
\bottomrule
\end{tabular}
\end{center}
\vspace{-15pt}\caption{Evaluation of state-of-the-art video-based methods on 3DPW \cite{Marcard2018RecoveringA3}, MPI-INF-3DHP \cite{Mehta2017Monocular3H}, and Human3.6M \cite{h36m_pami} datasets. Following Choi \etal \cite{choi2020beyond}, all methods are trained on the training set including 3DPW, but do not use the Human3.6M SMPL parameters obtained from Mosh \cite{10.1145/2661229.2661273}. The number of input frames follows the original protocol of each method.}\vspace{-11pt}
\label{tbl:table1}
\end{table*}

\vspace{-1pt}
\section{Implementation details}\vspace{-4pt}


Following the previous works \cite{Kocabas2020VIBEVI,choi2020beyond}, we set $T=16$ as the sequence length. 
We use ResNet-50 \cite{He2016DeepRL} pre-trained by Kolotouros \etal \cite{Kolotouros2019LearningTR} to serve as our static feature extractor. The static feature extractor is fixed and outputs a $2048$-dimensional feature for each frame, \ie  $\bx_t \in \bbR^{2048}$. The SMPL parameter regressor has two FC layers, each with $1024$ neurons, and followed an output layer to output $85$ pose, shape, and camera parameters $\Theta_t$ for each frame \cite{hmrKanazawa17, Kolotouros2019LearningTR}. The discriminator architecture we use is the same as \cite{Kocabas2020VIBEVI}. The parameters of MPS-Net and discriminator are optimized by the Adam solver \cite{Kingma2015AdamAM} at a learning rate of $5 \times 10^{-5}$ and $1 \times 10^{-4}$, respectively. The mini-batch size is set to $32$. During training, if the performance does not improve within $5$ epochs, the learning rate of both the MPS-Net and the discriminator will be reduced by a factor of $10$. We use an NVIDIA Titan RTX GPU to train the entire network for $30$ epochs. PyTorch \cite{pytorch} is used for code implementation.\vspace{-5pt}

\section{Experiments}\vspace{-4pt}

We first illustrate the datasets used for training and evaluation and the evaluation metrics. Then, we compare our MPS-Net against other state-of-the-art video-based methods and single image-based methods to demonstrate its advantages in addressing 3D human pose and shape estimation. We also provide an ablation study to confirm the effectiveness of each module in MPS-Net. Finally, we visualize some examples to show the qualitative evaluation results.

\vspace{5pt}\noindent{\bf Datasets.} Following the previous works \cite{Kocabas2020VIBEVI,choi2020beyond}, we adopt batches of mixed 3D and 2D datasets for training. For 3D datasets, we use 3DPW \cite{Marcard2018RecoveringA3}, MPI-INF-3DHP \cite{Mehta2017Monocular3H}, Human3.6M \cite{h36m_pami}, and AMASS \cite{Mahmood2019AMASSAO} for training, where 3DPW and AMASS provide SMPL parameter annotations, while MPI-INF-3DHP and Human3.6M include 3D joint annotations. For 2D datasets, we use PoseTrack \cite{Andriluka_2018_CVPR} and InstaVariety \cite{Kanazawa2019Learning3H} for training, where PoseTrack provides ground-truth 2D joints, while InstaVariety includes pseudo ground-truth 2D joints annotated using a 2D keypoint detector \cite{Cao2017RealtimeM2}. In terms of evaluation, the 3DPW, MPI-INF-3DHP, and Human3.6M datasets are used. Among them, Human3.6M is an indoor dataset, while 3DPW and MPI-INF-3DHP contain challenging outdoor videos.
More detailed settings are in the supplementary material.

\vspace{3pt}\noindent{\bf Evaluation metrics.} For the evaluation, four standard metrics are used \cite{Kocabas2020VIBEVI,choi2020beyond,Luo_2020_ACCV}, including the mean per joint position error (MPJPE), the Procrustes-aligned mean per joint position error (PA-MPJPE), the mean per vertex position error (MPVPE), and the acceleration error (ACC-ERR). Among them, MPJPE, PA-MPJPE, and MPVPE are mainly used to express the accuracy of the estimated 3D human pose and shape (measured in millimeter ($mm$)), and ACC-ERR ($mm/s^2$) is used to express the smoothness and temporal coherence of 3D human motion. A detailed description of each metric is included in the supplementary material.\vspace{-5pt}

\subsection{Comparison with state-of-the-art methods}

\noindent{\bf Video-based methods.} Table \ref{tbl:table1} shows the performance comparison between our MPS-Net and the state-of-the-art video-based methods on the 3DPW, MPI-INF-3DHP, and Human3.6M datasets. Following TCMR \cite{choi2020beyond}, all methods are trained on the training set including 3DPW, but do not use the Human3.6M SMPL parameters obtained from Mosh \cite{10.1145/2661229.2661273} for supervision. Because the SMPL parameters from Mosh have been removed from public access due to legal issues \cite{Luo_2020_ACCV}. The values of the comparison method are from TCMR \cite{choi2020beyond}, but we validated them independently.

\begin{table}
\fontsize{7.3pt}{10pt}\selectfont
\begin{center}
\begin{tabular}{l|ccc}
\toprule 
 & \#Parameters (M) & FLOPs (G) & Model Size (MB) \\
\midrule
VIBE \cite{Kocabas2020VIBEVI} & 72.43 & \bf{4.17} & 776 \\
\rowcolor{black!10} MEVA \cite{Luo_2020_ACCV} & 85.72 & 4.46 & 858.8 \\
TCMR \cite{choi2020beyond} & 108.89 & 4.99 & 1073 \\
\rowcolor{black!10} MPS-Net (Ours) &\bf{39.63} & 4.45 & \bf{331} \\
\bottomrule
\end{tabular}
\end{center}
\vspace{-18pt}\caption{Comparison of the number of network parameters, FLOPs, and model size.}\vspace{-8pt}
\label{tbl:table2}
\end{table}

\begin{table}
\fontsize{7.3pt}{10pt}\selectfont
\begin{center}
\begin{tabular}{ *{5}{l@{\hspace{0.15cm}}|c@{\hspace{0.15cm}}c@{\hspace{0.15cm}}c@{\hspace{0.15cm}}c@{\hspace{0.15cm}}} }
\toprule 
&\multicolumn{4}{c}{3DPW} \\
\cmidrule(lr){2-5}
Method & PA-MPJPE $\downarrow$ & MPJPE $\downarrow$ & MPVPE $\downarrow$ & ACC-ERR $\downarrow$ \\
\midrule
MPS-Net & \multirow{2}*{54.1} & \multirow{2}*{87.6} & \multirow{2}*{103.1} & \multirow{2}*{24.1} \\
- only Non-local \cite{Wang2018NonlocalNN} & & & & \\
\rowcolor{black!10}MPS-Net &  &  &  &  \\ 
\rowcolor{black!10}- only MoCA& \raisebox{.5\normalbaselineskip}[0pt][0pt]{53.0}& \raisebox{.5\normalbaselineskip}[0pt][0pt]{86.7} & \raisebox{.5\normalbaselineskip}[0pt][0pt]{102.2}& \raisebox{.5\normalbaselineskip}[0pt][0pt]{23.5}\\
MPS-Net & \multirow{2}*{52.4} & \multirow{2}*{86.0} & \multirow{2}*{101.5} & \multirow{2}*{10.5} \\
 - MoCA + TF-intgr. \cite{choi2020beyond} & & & & \\
\rowcolor{black!10}MPS-Net (Ours) &  &  &  & \\
\rowcolor{black!10} - MoCA + HAFI & \raisebox{.5\normalbaselineskip}[0pt][0pt]{\bf{52.1}} & \raisebox{.5\normalbaselineskip}[0pt][0pt]{\bf{84.3}}& \raisebox{.5\normalbaselineskip}[0pt][0pt]{\bf{99.7}}& \raisebox{.5\normalbaselineskip}[0pt][0pt]{\bf{7.4}}\\
\bottomrule
\end{tabular}
\end{center}
\vspace{-18pt}\caption{Ablation study for different modules of the MPS-Net on the 3DPW \cite{Marcard2018RecoveringA3} dataset. The training and evaluation settings are the same as the experiments on the 3DPW dataset in Table \ref{tbl:table1}.}\vspace{-6pt}
\label{tbl:table3}
\end{table}

\begin{table}
\fontsize{7.3pt}{10pt}\selectfont
\begin{center}
\begin{tabular}{c@{\hspace{0.04cm}}c@{\hspace{0.24cm}}l@{\hspace{0.12cm}} | c@{\hspace{0.12cm}}c@{\hspace{0.12cm}}c@{\hspace{0.12cm}}c@{\hspace{0.12cm}}}
\toprule 
& & &\multicolumn{4}{c}{3DPW}\\
\cmidrule(lr){4-7} 
 & & Method & PA-MPJPE $\downarrow$ & MPJPE $\downarrow$ & MPVPE $\downarrow$ & ACC-ERR $\downarrow$\\
\midrule
\multirow{5}{*}{\rotatebox{90}{single image}} & \multirow{5}{*}{\rotatebox{90}{-based}}& HMR \cite{hmrKanazawa17} & 76.7 & 130.0 & - & 37.4 \\
 & &\cellcolor{black!10}GraphCMR \cite{Kolotouros2019ConvolutionalMR} & \cellcolor{black!10}70.2 & \cellcolor{black!10}- & \cellcolor{black!10}- & \cellcolor{black!10}- \\ 
 & &SPIN \cite{Kolotouros2019LearningTR} & 59.2 & 96.9 & 116.4 & 29.8 \\
 & &\cellcolor{black!10}PyMAF \cite{pymaf2021} & \cellcolor{black!10}58.9 & \cellcolor{black!10}92.8 & \cellcolor{black!10}110.1 & \cellcolor{black!10}- \\
 & &I2L-MeshNet \cite{Moon_2020_ECCV_I2L-MeshNet} & 57.7 & 93.2 & 110.1 & 30.9 \\
\midrule
\multirow{6}{*}{\rotatebox{90}{video-based}} & & \cellcolor{black!10}HMMR \cite{Kanazawa2019Learning3H} & \cellcolor{black!10}72.6 & \cellcolor{black!10}116.5 & \cellcolor{black!10}139.3 & \cellcolor{black!10}15.2 \\
 & & Doersch \etal \cite{Doersch2019Sim2realTL} & 74.7 & - & - & - \\
 & & \cellcolor{black!10}Sun \etal \cite{Yu2019HumanMR} & \cellcolor{black!10}69.5 & \cellcolor{black!10}- & \cellcolor{black!10}- & \cellcolor{black!10}- \\
 & & VIBE \cite{Kocabas2020VIBEVI} & 56.5 & 93.5 & 113.4 & 27.1 \\
 & & \cellcolor{black!10}TCMR \cite{choi2020beyond} & \cellcolor{black!10}55.8 & \cellcolor{black!10}95.0 & \cellcolor{black!10}111.3 & \cellcolor{black!10}\bf{6.7} \\
 & & MPS-Net (Ours) & \bf{54.0} & \bf{91.6} & \bf{109.6} & 7.5 \\
\bottomrule
\end{tabular}
\end{center}
\vspace{-18pt}\caption{Evaluation of state-of-the-art single image-based and video-based methods on the 3DPW \cite{Marcard2018RecoveringA3} dataset. All methods do not use 3DPW for training.}\vspace{-18pt}
\label{tbl:table4}
\end{table}

The results in Table \ref{tbl:table1} show that our MPS-Net outperforms the existing video-based methods in almost all metrics and datasets. This demonstrates that by capturing the motion continuity dependencies and integrating temporal features from adjacent past and future, performance can indeed be improved. Although TCMR \cite{choi2020beyond} has also made great progress, it is limited by the ability of recurrent operation (\ie GRU) to capture non-local context relations in the action sequence \cite{Vaswani2017AttentionIA,Wang2018NonlocalNN}, thereby reducing the accuracy of the estimated 3D human pose and shape (\ie PA-MPJPE, MPJPE, and MPVPE are higher than MPS-Net). In addition, the number of network parameters and model size of TCMR are also about $3$ times that of MPS-Net (see Table \ref{tbl:table2}), which is relatively heavy. Regarding MEVA \cite{Luo_2020_ACCV}, as shown in Table \ref{tbl:table1}, MEVA requires at least $90$ input frames, which means it cannot be trained and tested on short videos. This greatly reduces the value in practical applications. Overall, our MPS-Net can effectively estimate accurate (lower PA-MPJPE, MPJPE, and MPVPE) and smooth (lower ACC-ERR) 3D human pose and shape from a video, and is relatively lightweight (fewer network parameters). The comparisons on the three datasets also show the strong generalization property of our MPS-Net.

\vspace{1pt}
\noindent{\bf Ablation analysis.} 
To analyze the effectiveness of the MoCA and HAFI modules in MPS-Net, we conduct ablation studies on MPS-Net under the challenging in-the-wild 3DPW dataset. Specifically, we evaluate the impact on MPS-Net by replacing the MoCA module with the non-local block \cite{Wang2018NonlocalNN}, considering only the MoCA module (without using HAFI), and replacing the HAFI module with the temporal feature integration scheme proposed by Choi \etal \cite{choi2020beyond}. For performance comparison, it is obvious from Table \ref{tbl:table3} that the proposed MoCA module (\ie MPS-Net-only MoCA) is superior to non-local block (\ie MPS-Net-only Non-local) in all metrics. The results confirm that by further introducing the a \emph{priori} knowledge of NSSM to guide self-attention learning, the MoCA module can indeed improve 3D human pose and shape estimation. On the other hand, the results also show that our HAFI module (\ie MPS-Net-MoCA+HAFI) outperforms the temporal feature integration scheme (\ie MPS-Net-MoCA+TF-intgr.), which demonstrates that the gradual integration of adjacent features through a hierarchical attentive integration manner can indeed strengthen temporal correlation and make the generated 3D human motion smoother (\ie lower ACC-ERR). 
Overall, the ablation analysis confirmed the effectiveness of the proposed MoCA and HAFI modules.\vspace{1pt}

\noindent{\bf Single image-based and video-based methods.} We further compare our MPS-Net with the methods including single image-based methods on the challenging in-the-wild 3DPW dataset. Notice that a number of previous works \cite{hmrKanazawa17,Kolotouros2019ConvolutionalMR,Kolotouros2019LearningTR,pymaf2021,Moon_2020_ECCV_I2L-MeshNet,Kanazawa2019Learning3H,Doersch2019Sim2realTL,Yu2019HumanMR,Kocabas2020VIBEVI,choi2020beyond} did not use the 3DPW training set to train their models, so in the comparison in Table \ref{tbl:table4}, all methods are not trained on 3DPW.\vspace{-2pt}

\begin{figure}[t]
  \centering
  \animategraphics[autoplay,
                    loop,
                    width=\linewidth]{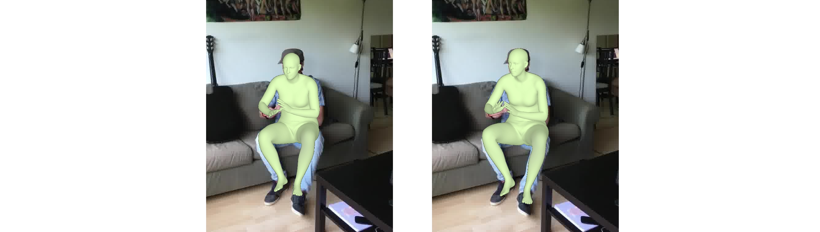}{./Fig6gif2png/}{1}{361}\vspace{-8pt}
  \caption{Qualitative comparison of TCMR \cite{choi2020beyond} (left) and our MPS-Net (right) on the challenging in-the-wild 3DPW \cite{Marcard2018RecoveringA3} dataset (the 1st and 2nd clips) and MPI-INF-3DHP \cite{Mehta2017Monocular3H} dataset (the 3rd clip). \textit{This is an embedded video, please use Adobe Acrobat (standalone version) to view it. Note that the video animation has been slowed down for better viewing.}}\vspace{-2pt}
  \label{fig:fig6}
\end{figure}

\begin{figure}
  \centering
  \includegraphics[width=\linewidth]{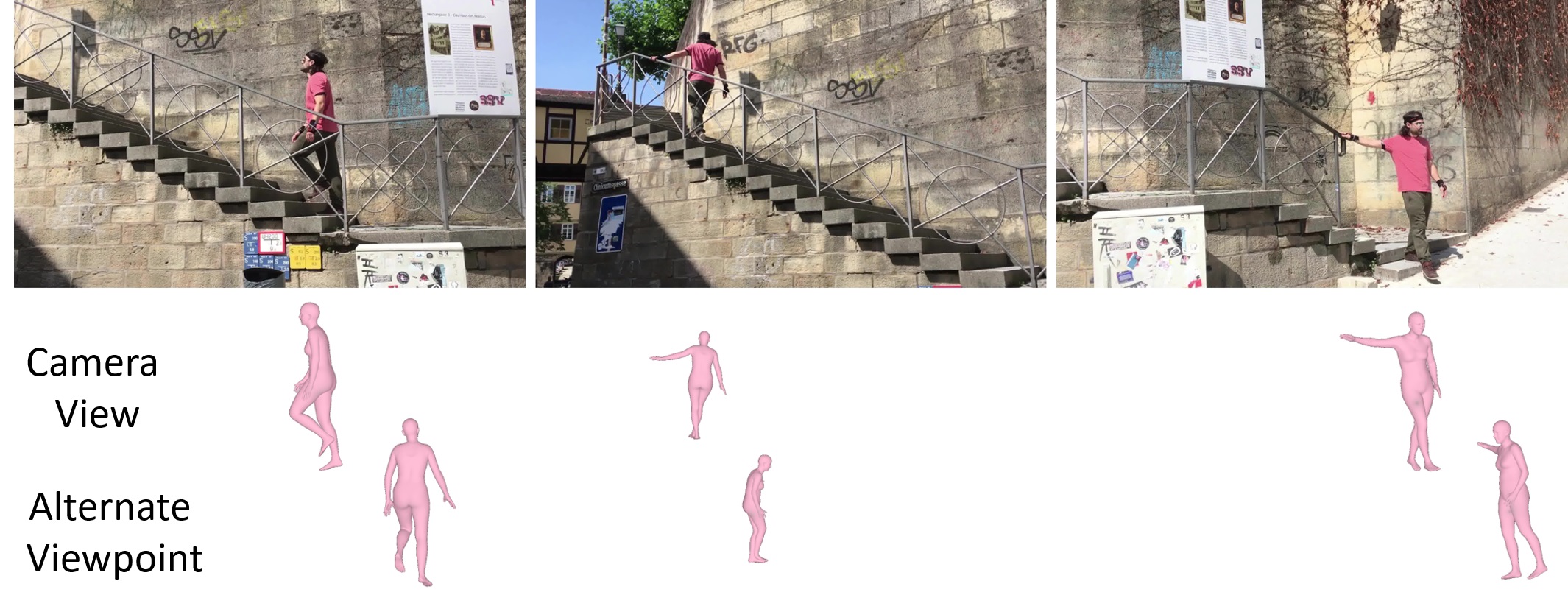}\\
  \includegraphics[width=\linewidth]{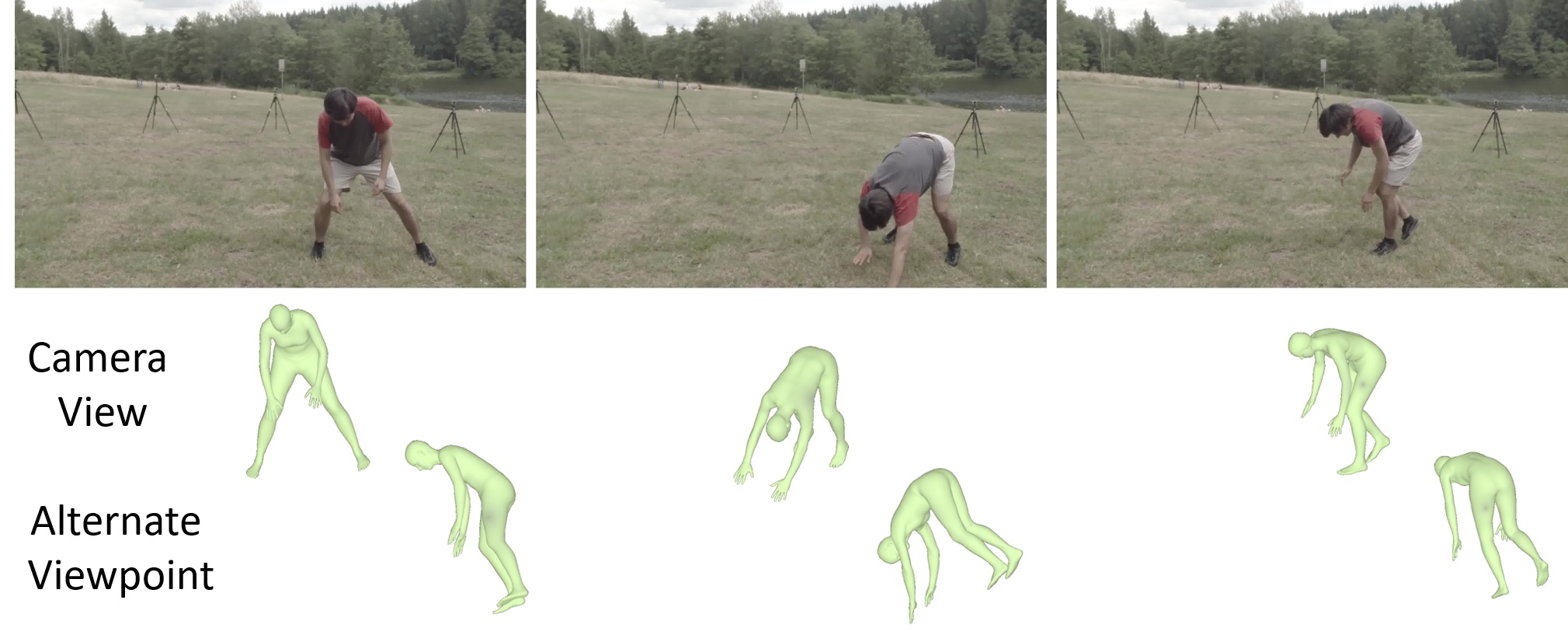}\vspace{-8pt}
  \caption{Qualitative results of MPS-Net on the challenging in-the-wild 3DPW \cite{Marcard2018RecoveringA3} dataset and MPI-INF-3DHP \cite{Mehta2017Monocular3H} dataset. For each sequence, the top row shows input images, the middle row shows the estimated body mesh from the camera view, and the bottom row shows the estimated mesh from an alternate viewpoint.}\vspace{-12pt}
  \label{fig:fig7}
\end{figure}

Similar to the results in Table \ref{tbl:table1}, the results in Table \ref{tbl:table4} demonstrate that our MPS-Net performs favorably against existing single image-based and video-based methods on the PA-MPJPE, MPJPE, and MPVPE evaluation metrics. Although TCMR achieves the lowest ACC-ERR, it tends to be overly smooth, thereby sacrificing the accuracy of pose and shape estimation. Specifically, when TCMR reduces ACC-ERR $0.8$ $mm/s^2$ compared to MPS-Net, MPS-Net reduces PA-MPJPE, MPJPE, and MPVPE by $1.8$ $mm$, $3.4$ $mm$, and $1.7$ $mm$, respectively. Table \ref{tbl:table4} further confirms the importance of considering temporal information in consecutive frames, \ie compared with single-image-based methods, video-based methods have lower ACC-ERR. In summary, MPS-Net achieves a better balance in the accuracy and smoothness of 3D human pose and shape estimation.\vspace{-4pt}

\subsection{Qualitative evaluation}\vspace{-4pt}


We present 1) visual comparisons with the TCMR \cite{choi2020beyond}, 2) visual effects of MPS-Net in alternative viewpoints, and 3) visual results of the learned human motion continuity.\vspace{2pt}

\noindent{\bf Visual comparisons with the TCMR.} The qualitative comparison between TCMR and our MPS-Net on the 3DPW and MPI-INF-3DHP datasets is shown in Figure \ref{fig:fig6}. From the results, we observe that the 3D human pose and shape estimated by MPS-Net can fit the input images well, especially on the limbs. TCMR seems to be too focused on generating smooth 3D human motion, so the estimated pose has relatively small changes from frame to frame, which limits its ability to fit the input images.\vspace{3pt}

\noindent{\bf Visual effects of MPS-Net in alternative viewpoints.} We visualize the 3D human body estimated by MPS-Net from different viewpoints in Figure \ref{fig:fig7}. The results show that MPS-Net can estimate the correct global body rotation. 
This is quantitatively demonstrated by the improvements in the PA-MPJPE, MPJPE, and MPVPE (see Table \ref{tbl:table1}).\vspace{1pt}



\begin{figure}[t]
  \centering
  \includegraphics[width=\linewidth]{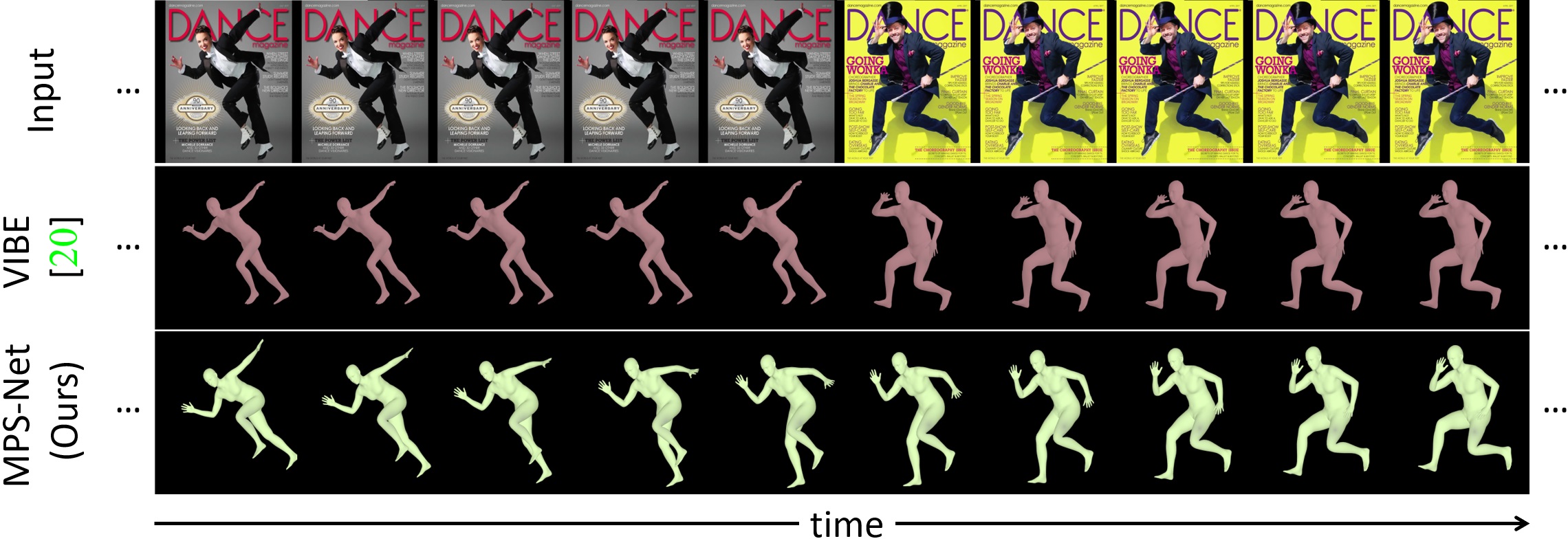}\vspace{-10pt}
  \caption{An example of visualization of the VIBE \cite{Kocabas2020VIBEVI} and our MPS-Net on the continuity of human motion.}\vspace{-13pt}
  \label{fig:fig8}
\end{figure}

\noindent{\bf Visual results of the learned human motion continuity.} We use a relatively extreme example to show the continuity of human motion learned by MPS-Net. In this example, we randomly downloaded two pictures with different poses from the Internet, and copied the pictures multiple times to form a sequence. Then, we send the sequence to VIBE \cite{Kocabas2020VIBEVI} and MPS-Net for 3D human pose and shape estimation. As shown in Figure \ref{fig:fig8}, compared with VIBE, it is obvious from the estimation results that our MPS-Net produces a transition effect between pose exchanges, and this transition conforms to the continuity of human kinematics. It demonstrates that MPS-Net has indeed learned the continuity of human motion, and explains why MPS-Net can achieve lower ACC-ERR in the benchmark (action) datasets (see Table \ref{tbl:table1}). This result is also similar to using a 3D motion predictor to estimate reasonable human motion in-betweening of two key frames \cite{10.1145/3386569.3392480}. 
In contrast, VIBE relies too much on the features of the current frame, making it unable to truly learn the continuity of human motion. Thus, its ACC-ERR is still high (see Table \ref{tbl:table1}).

For more results and video demos can be found at \url{https://mps-net.github.io/MPS-Net/}.\vspace{-6pt}

\vspace{-4pt}
\section{Conclusion}\vspace{-7pt}

We propose the MPS-Net for estimating 3D human pose and shape from monocular video. The main contributions of this work lie in the design of the MoCA and HAFI modules. The former leverages visual cues observed from human motion to adaptively recalibrate the range that needs attention in the sequence to capture the motion continuity dependencies, and the later allows our model to strengthen temporal correlation and refine feature representation for producing temporally coherent estimates. Compared with existing methods, the integration of MoCA and HAFI modules demonstrates the advantages of our MPS-Net in achieving the state-of-the-art 3D human pose and shape estimation.\vspace{2pt}


\noindent \textbf{Acknowledgment:} 
This work was supported in part by MOST under grants 110-2221-E-001-016-MY3, 110-2634-F-007-027 and 110-2634-F-002-050, and Academia Sinica under grant AS-TP-111-M02.



{\small
\bibliographystyle{ieee_fullname}
\bibliography{egbib}
}

\clearpage

\pagebreak
\begin{center}
\textbf{\large Supplementary Material}
\end{center}
\makeatletter

\section{Datasets}

\vspace{5pt}\noindent{\bf 3DPW.} 3DPW \cite{Marcard2018RecoveringA3} is mainly captured from outdoors and in-the-wild. It combines a hand-held camera and a set of inertial measurement unit (IMU) sensors attached at the body limbs to calculate the near ground-truth SMPL parameters. It contains a total of $60$ videos of different lengths. We use the official split to train and test the model, where the training, validation, and test sets are composed of $24$, $12$, and $24$ videos, respectively. In addition, we report MPVPE on 3DPW because it is the only dataset that contains ground-truth 3D shape annotations among the datasets we used.

\vspace{5pt}\noindent{\bf MPI-INF-3DHP.} MPI-INF-3DHP \cite{Mehta2017Monocular3H} is a dataset consisting of both constrained indoor and complex outdoor scenes. It is captured using a multi-view camera setting with a markerless motion capture system, and the 3D joint annotation is calculated through the multiview method. Following existing methods \cite{Kocabas2020VIBEVI,choi2020beyond}, we use the official split to train and test the model. The training set contains $8$ subjects, each of which has $16$ videos, and the test set contains $6$ subjects, performing $7$ actions in both indoor and outdoor environments.

\vspace{5pt}\noindent{\bf Human3.6M.} Human3.6M \cite{h36m_pami} is one of the largest motion capture datasets, which contains $3.6$ million video frames and corresponding 3D joint annotations. This dataset is collected in an indoor and controlled environment. Same as existing methods \cite{Kocabas2020VIBEVI,choi2020beyond}, we train the model on $5$ subjects (\ie S1, S5, S6, S7, and S8) and evaluate it on $2$ subjects (\ie S9 and S11). We subsampled the dataset to $25$ frames per second (fps) for training and testing.

\vspace{5pt}\noindent{\bf AMASS.} AMASS \cite{Mahmood2019AMASSAO} is a large-scale human motion sequence database that unifies $15$ existing motion capture (mocap) datasets by representing them within a common framework and parameterization. These motion sequences are annotated with Mosh++ to generate SMPL parameters. AMASS has a total of $42$ hours of mocap, $346$ subjects, and $11,451$ human motions. Following the setting of the existing method \cite{Kocabas2020VIBEVI}, we use this database to train our MPS-Net.

\vspace{5pt}\noindent{\bf PoseTrack.} PoseTrack \cite{Andriluka_2018_CVPR} is a 2D benchmark dataset for video-based multi-person pose estimation and articulated tracking. It contains a total of $1,337$ videos, divided into $792$, $170$, and $375$ videos for training, validation, and testing. Each person instance in the video is annotated with $15$ keypoints. Same as existing methods \cite{Kocabas2020VIBEVI,choi2020beyond}, we use the training set for model training.

\vspace{5pt}\noindent{\bf InstaVariety.} InstaVariety \cite{Kanazawa2019Learning3H} is a 2D benchmark dataset captured from Instagram using 84 hashtags. There are $28,272$ videos in total, with an average length of $6$ seconds, and OpenPose \cite{Cao2017RealtimeM2} is used to acquire pseudo ground-truth 2D joint annotations. Same as existing methods \cite{Kocabas2020VIBEVI,choi2020beyond}, we adopt this dataset for training.

\section{Evaluation metrics}

Four standard evaluation metrics \cite{Kocabas2020VIBEVI,choi2020beyond,Luo_2020_ACCV} are considered, including MPJPE, PA-MPJPE, MPVPE, and ACC-ERR. Specifically, MPJPE is calculated as the mean of the Euclidean distance between the ground-truth and the estimated 3D joint positions after aligning the pelvis joint on the ground truth location. PA-MPJPE is calculated similarly to MPJPE, but after the estimated pose is rigidly aligned with the ground-truth pose. MPVPE is calculated as the mean of the Euclidean distance between the ground truth and the estimated 3D mesh vertices (output by the SMPL model). ACC-ERR is measured as the mean difference between the ground-truth and the estimated 3D acceleration for every joint.

\section{Ablation study of HAFI module}

To analyze the effect of the number of frames per group in the HAFI module, we conduct ablation studies on MPS-Net with different HAFI module settings under the 3DPW dataset \cite{Marcard2018RecoveringA3}. The results in Table \ref{tbl:table5} indicate that considering the temporal features of three adjacent frames as a group can enable our MPS-Net to achieve the best 3D human pose and shape estimation. Therefore, for all experiments (\ie Table \ref{tbl:table1} to Table \ref{tbl:table4}), the HAFI module defaults to three frames as a group. 

\begin{table}
\fontsize{7.3pt}{10pt}\selectfont
\begin{center}
\begin{tabular}{ *{5}{l@{\hspace{0.15cm}}|c@{\hspace{0.15cm}}c@{\hspace{0.15cm}}c@{\hspace{0.15cm}}c@{\hspace{0.15cm}}} }
\toprule 
&\multicolumn{4}{c}{3DPW} \\
\cmidrule(lr){2-5}
Method & PA-MPJPE $\downarrow$ & MPJPE $\downarrow$ & MPVPE $\downarrow$ & ACC-ERR $\downarrow$ \\
\midrule
MPS-Net & \multirow{2}*{52.6} & \multirow{2}*{85.4} & \multirow{2}*{101.0} & \multirow{2}*{7.8} \\
(HAFI, 2 frames/group) & & & & \\
\rowcolor{black!10}MPS-Net &  &  &  &  \\ 
\rowcolor{black!10}(HAFI, 3 frames/group)& \raisebox{.5\normalbaselineskip}[0pt][0pt]{\bf{52.1}}& \raisebox{.5\normalbaselineskip}[0pt][0pt]{\bf{84.3}} & \raisebox{.5\normalbaselineskip}[0pt][0pt]{\bf{99.7}}& \raisebox{.5\normalbaselineskip}[0pt][0pt]{\bf{7.4}}\\
MPS-Net & \multirow{2}*{52.5} & \multirow{2}*{85.9} & \multirow{2}*{101.2} & \multirow{2}*{7.6} \\
(HAFI, 4 frames/group) & & & & \\
\bottomrule
\end{tabular}
\end{center}
\vspace{-18pt}\caption{Effect of the number of frames per group in the HAFI module. The training and evaluation settings are the same as the experiments on the 3DPW dataset \cite{Marcard2018RecoveringA3} in Table \ref{tbl:table1}.}
\label{tbl:table5}
\end{table}

On the other hand, we further add the results considering only the HAFI module on MPS-Net (called MPS-Net-only HAFI) to supplement the ablation experiments in Table \ref{tbl:table3}, with results of 54.0, 87.6, 103.5, and 7.5, respectively, for PA-MPJPE, MPJPE, MPVPE and ACC-ERR. Compared with MPS-Net-only MoCA (see Table \ref{tbl:table3}), MPS-Net-only HAFI yields better ACC-ERR but worse on PA-MPJPE, MPJPE and MPVPE. However, by coupling MoCA with HAFI, MPS-Net (Ours) achieves the best result.


\begin{figure*}[t]
  \centering
  \includegraphics[width=\linewidth]{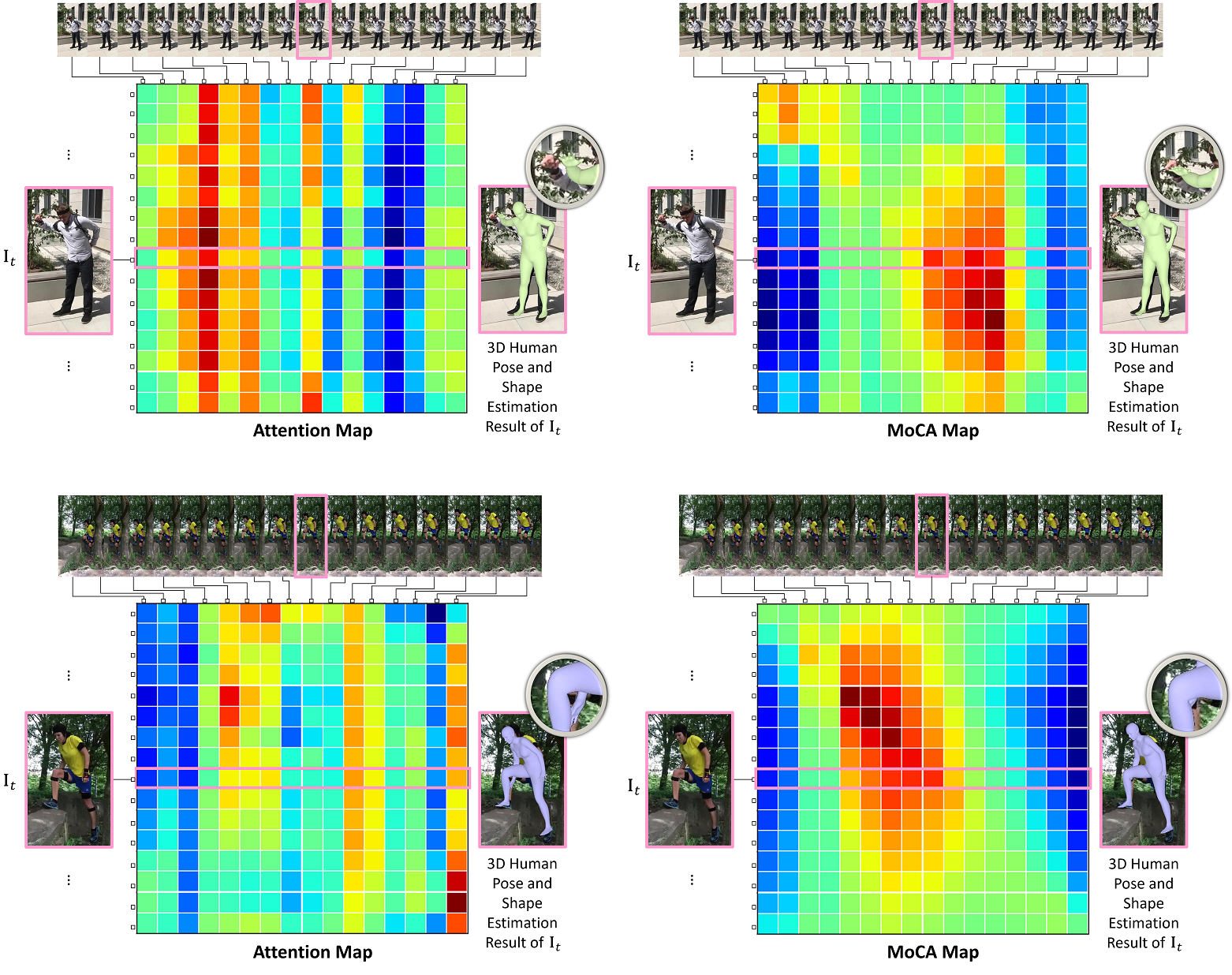}
  \caption{Visual comparison of 3D human pose and shape estimation of MoCA module and non-local block \cite{Wang2018NonlocalNN} on the 3DPW dataset \cite{Marcard2018RecoveringA3}. Where the attention map is generated from the non-local operation, and the MoCA map is generated from the MoCA operation. In the attention and MoCA maps, red indicates a higher attention value, and blue indicates a lower one. The results demonstrate that the MoCA map generated by our MoCA operation can indeed allow the MPS-Net to focus attention on a more appropriate range of action sequence to improve the estimation results.}
  \label{fig:fig9}
\end{figure*}

\section{Qualitative comparison between MoCA module and non-local block}

To further verify whether the proposed MoCA operation can improve the 3D human pose and shape estimation by introducing NSSM to recalibrate the attention map generated by the non-local operation \cite{Wang2018NonlocalNN}, we conduct the following qualitative experiments. Specifically, we visualize the 3D human pose and shape estimation resulted from the methods of MPS-Net-only Non-local and MPS-Net-only MoCA in Table \ref{tbl:table3}, respectively. The results can be seen from the two examples in Figure \ref{fig:fig9} that the MoCA map generated by our MoCA operation can indeed allow the MPS-Net to focus attention on a more appropriate range of action sequence, thereby improving the accuracy of 3D human pose and shape estimation. On the contrary, the attention map generated by non-local operation is often unstable, and it is easy to focus attention on less correlated frames and ignore the continuity of human motion in the action sequence, which reduces the accuracy of estimation. Such a result is quantitatively demonstrated by the improvement of MPS-Net-only MoCA in the MPJPE, PA-MPJPE, and MPVPE errors (see Table \ref{tbl:table3}).

On the other hand, we also add the results considering only the setting of NSSM on MPS-Net (called MPS-Net-only NSSM) to supplement the ablation experiments in Table \ref{tbl:table3}, with results of 53.3, 88.0, 104.1, and 26.0, respectively, for PA-MPJPE, MPJPE, MPVPE and ACC-ERR. These (cf. Table \ref{tbl:table3}) indicate that MPS-Net-only NSSM and MPS-Net-only Non-local are complementary, and their fusion, \ie MPS-Net-only MoCA, can further achieve better results.

\section{Effect of motion speed and input fps}


To demonstrate how motion speed and frame rate influence MPS-Net performance, we conducted experiments with various demo videos (\eg fast-moving dancing or general walking video) at different frame rates, from 24fps to 60fps, and found that it has little impact on MPS-Net. Frame rate info for each demo video is available on our demo website.

\end{document}